\documentclass{article}
\usepackage{ijcai25}

\usepackage{times}
\usepackage{soul}
\usepackage{url}
\usepackage[hidelinks]{hyperref}
\usepackage[utf8]{inputenc}
\usepackage[small]{caption}
\usepackage{graphicx}
\usepackage{amsmath}
\usepackage{amsthm}
\usepackage{booktabs}
\usepackage{algorithm}
\usepackage{algorithmic}
\usepackage[switch]{lineno}

\usepackage{amsmath,amssymb,amsfonts}
\usepackage{algorithmic}
\usepackage{graphicx}
\usepackage{textcomp}
\usepackage{xcolor}

\usepackage{bbold}
\usepackage{graphicx}
\usepackage{multirow}
\usepackage{subfigure}
\usepackage{enumitem}
\usepackage{soul}
\usepackage{color}
\usepackage{acronym}
\usepackage{threeparttable}
\usepackage{url}
\usepackage{epsfig}
\usepackage{float}
\usepackage[usestackEOL]{stackengine}

\usepackage{hyperref}

\definecolor{ForestGreen}{RGB}{34,139,34}

\newcommand{\red}[1]{\textcolor{red}{#1}}
\newcommand{\blue}[1]{\textcolor{blue}{#1}}

\usepackage[font=bf]{caption}
\usepackage[T1]{fontenc}
\usepackage{fancyvrb}

\usepackage{xspace}

\newcommand{\eg}{{\it e.g.}}

\newcommand{\etc}{{\it etc}.~}
\newcommand{\ie}{{\it i.e.}}

\newcommand{\sys}{\textsc{THOR}\xspace}
\title{\sys: A Generic Energy Estimation Approach for On-Device Training}

\iftrue
\author{
Jiaru Zhang$^1$
\and
Zesong Wang$^1$\and
Hao Wang$^2$ \and
Tao Song$^1$\footnote{Corresponding author.} \and
Huai-an Su$^3$ \and
Rui Chen$^3$ \and \\
Yang Hua$^4$ \and
Xiangwei Zhou$^5$ \and
Ruhui Ma$^1$ \and
Miao Pan$^3$ \And
Haibing Guan$^1$\\
\affiliations
$^1$Shanghai Jiao Tong University
$^2$Stevens Institute of Technology\\
$^3$University of Houston
$^4$Queen's University Belfast
$^5$Louisiana State University\\
\emails
\{jiaruzhang, taihaozesong, songt333, ruhuima, hbguan\}@sjtu.edu.cn, \\
hwang9@stevens.edu,
\{hsu3, rchen19, mpan2\}@uh.edu,
Y.Hua@qub.ac.uk, xwzhou@lsu.edu
}
\fi

\begin{document}

\acrodef{ML}[ML]{Machine Learning}
\newcommand{\ML}{\ac{ML}\xspace}

\acrodef{NSF}[NSF]{National Science Foundation}
\newcommand{\NSF}{\ac{NSF}\xspace}

\acrodef{AI}[AI]{Artificial Intelligence}
\newcommand{\AI}{\ac{AI}\xspace}

\acrodef{FL}[FL]{Federated Learning}
\newcommand{\FL}{\ac{FL}\xspace}

\acrodef{DVFS}[DVFS]{Dynamic Voltage and Frequency Scaling}
\newcommand{\DVFS}{\ac{DVFS}\xspace}

\acrodef{GP}[GP]{Gaussian Process}
\newcommand{\GP}{\ac{GP}\xspace}

\acrodef{RBF}[RBF]{Radial Basis Function}
\newcommand{\RBF}{\ac{RBF}\xspace}

\acrodef{GPU}[GPU]{Graphics Processing Unit}
\acrodefplural{GPU}[GPUs]{Graphics Processing Units}
\newcommand{\GPU}{\ac{GPU}\xspace}
\newcommand{\GPUs}{\acp{GPU}\xspace}

\acrodef{NLP}[NLP]{Natural Language Processing}
\newcommand{\NLP}{\ac{NLP}\xspace}

\acrodef{HAR}[HAR]{Human Activity Recognition}
\newcommand{\HAR}{\ac{HAR}\xspace}

\acrodef{CUDA}[CUDA]{Compute Unified Device Architecture}
\newcommand{\CUDA}{\ac{CUDA}\xspace}

\acrodef{BO}[BO]{Bayesian Optimization}
\newcommand{\BO}{\ac{BO}\xspace}

\acrodef{SGD}[SGD]{Stochastic Gradient Descent}
\newcommand{\SGD}{\ac{SGD}\xspace}

\acrodef{CDF}[CDF]{Cumulative Distribution Function}
\newcommand{\CDF}{\ac{CDF}\xspace}

\acrodef{MAPE}[MAPE]{Mean Absolute Percentage Error}
\newcommand{\MAPE}{\ac{MAPE}\xspace}

\acrodef{FLOP}[FLOP]{Floating Point Operation}
\acrodefplural{FLOP}[FLOPs]{Floating Point Operations}
\newcommand{\FLOP}{\ac{FLOP}\xspace}
\newcommand{\FLOPs}{\acp{FLOP}\xspace}

\acrodef{FLOPS}[FLOPS]{Floating Point Operation per Second}
\newcommand{\FLOPS}{\ac{FLOPS}\xspace}

\acrodef{FC}[FC]{Fully Connected}
\newcommand{\FC}{\ac{FC}\xspace}

\acrodef{CL}[CL]{Critical Learning}
\newcommand{\CL}{\ac{CL}\xspace}

\acrodef{AC}[AC]{Attacking-Critical}
\newcommand{\AC}{\ac{AC}\xspace}

\acrodef{CAGR}[CAGR]{compound annual growth rate}
\newcommand{\CAGR}{\ac{CAGR}\xspace}

\acrodef{CCT}[CCT]{Center for Computation and Technology}
\newcommand{\CCT}{\ac{CCT}\xspace}

\acrodef{SLO}[SLO]{service level objective}
\newcommand{\SLO}{\ac{SLO}\xspace}

\acrodefplural{SLOs}[SLO]{service level objectives}
\newcommand{\SLOs}{\acp{SLO}\xspace}

\acrodef{RL}[RL]{reinforcement learning}
\newcommand{\RL}{\ac{RL}\xspace}

\acrodef{DRL}[DRL]{deep reinforcement learning}
\newcommand{\DRL}{\ac{DRL}\xspace}

\acrodef{VM}[VM]{virtual machine}
\newcommand{\VM}{\ac{VM}\xspace}

\acrodefplural{VM}[VMs]{virtual machines}
\newcommand{\VMs}{\acp{VM}\xspace}

\acrodef{ITC}[ITC]{Innovation \& Technology Commercialization}
\newcommand{\ITC}{\ac{ITC}\xspace}

\acrodef{DAG}[DAG]{directed acyclic graph}
\newcommand{\DAG}{\ac{DAG}\xspace}

\acrodefplural{DAG}[DAGs]{directed acyclic graphs}
\newcommand{\DAGs}{\acp{DAG}\xspace}

\acrodef{SFA}[SFA]{single point authentication}
\newcommand{\SFA}{\ac{SFA}\xspace}

\acrodef{HPC}[HPC]{high-performance computing}
\newcommand{\HPC}{\ac{HPC}\xspace}

\acrodef{SBIR}[SBIR]{Small Business Innovation Research}
\newcommand{\SBIR}{\ac{SBIR}\xspace}

\acrodef{IoT}[IoT]{Internet of Things}
\newcommand{\IoT}{\ac{IoT}\xspace}

\acrodef{DML}[DML]{distributed machine learning}
\newcommand{\DML}{\ac{DML}\xspace}

\acrodef{GNN}[GNN]{graph neural network}
\newcommand{\GNN}{\ac{GNN}\xspace}

\acrodefplural{GNN}[GNNs]{graph neural networks}
\newcommand{\GNNs}{\acp{GNN}\xspace}

\acrodef{BSR}[BSR]{backdoor success rate}
\newcommand{\BSR}{\ac{BSR}\xspace}

\acrodef{BTA}[BTA]{backdoor task accuracy}
\newcommand{\BTA}{\ac{BTA}\xspace}

\acrodef{ATT}[ATT]{App Tracking Transparency}
\newcommand{\ATT}{\ac{ATT}\xspace}

\acrodef{DNN}[DNN]{deep neural network}
\newcommand{\DNN}{\ac{DNN}\xspace}

\acrodef{FA}[FA]{Federated Analytics}
\newcommand{\FA}{\ac{FA}\xspace}

\acrodef{PCA}[PCA]{Principal Component Analysis}
\newcommand{\PCA}{\ac{PCA}\xspace}

\acrodef{DP}[DP]{Differential Privacy}
\newcommand{\DP}{\ac{DP}\xspace}

\acrodef{DL}[DL]{Deep Learning}
\newcommand{\DL}{\ac{DL}\xspace}

\acrodef{CNN}[CNN]{Convolutional Neural Network}
\newcommand{\CNN}{\ac{CNN}\xspace}

\acrodef{LSTM}[LSTM]{Long Short Term Memory}
\newcommand{\LSTM}{\ac{LSTM}\xspace}

\acrodef{RNN}[RNN]{Recurrent Neural Network}
\newcommand{\RNN}{\ac{RNN}\xspace}

\acrodef{EHDEN}{European Health Data \& Evidence Network}
\newcommand{\EHDEN}{\ac{EHDEN}\xspace}

\acrodef{HE}{Homomorphic Encryption}
\newcommand{\HE}{\ac{HE}\xspace}

\acrodef{CPN}{Conditional Policy Network}
\newcommand{\CPN}{\ac{CPN}\xspace}
 
\acrodef{GDPR}[GDPR]{General Data Protection Regulation}
\newcommand{\GDPR}{\ac{GDPR}\xspace}

\maketitle

\begin{abstract} 
Battery-powered mobile devices (e.g., smartphones, AR/VR glasses, and various IoT devices) are increasingly being used for AI training due to their growing computational power and easy access to valuable, diverse, and real-time data.
On-device training is highly energy-intensive, making accurate energy consumption estimation crucial for effective job scheduling and sustainable AI. 
However, the heterogeneity of devices and the complexity of models challenge the accuracy and generalizability of existing methods.
This paper proposes \sys, a generic approach for energy consumption estimation in \DNN training. 
First, we examine the layer-wise energy additivity property of \acp{DNN} and strategically partition the entire model into layers for fine-grained energy consumption profiling. 
Then, we fit \GP models to learn from layer-wise energy consumption measurements and estimate a \DNN's overall energy consumption based on its layer-wise energy additivity property. 
We conduct extensive experiments with various types of models across different real-world platforms. 
The results demonstrate that \sys has effectively reduced the \MAPE by up to 30\%. 
Moreover, \sys is applied in guiding energy-aware pruning, successfully reducing energy consumption by 50\%, thereby further demonstrating its generality and potential.
\end{abstract}

\section{Introduction}
\label{sec:intro}

\AI applications are shifting from \textit{wall-plug-powered \AI devices} to \textit{battery-powered mobile AI systems},
such as smartphones, tablets, laptops, AR/VR wearables, and IoT devices.
Those mobile AI systems/devices are becoming increasingly powerful and provide easy access to a rich source of diverse, real-time data, making them highly valuable for training AI models.

\begin{figure}[t]
    \centering
    \includegraphics[width=.85\linewidth]{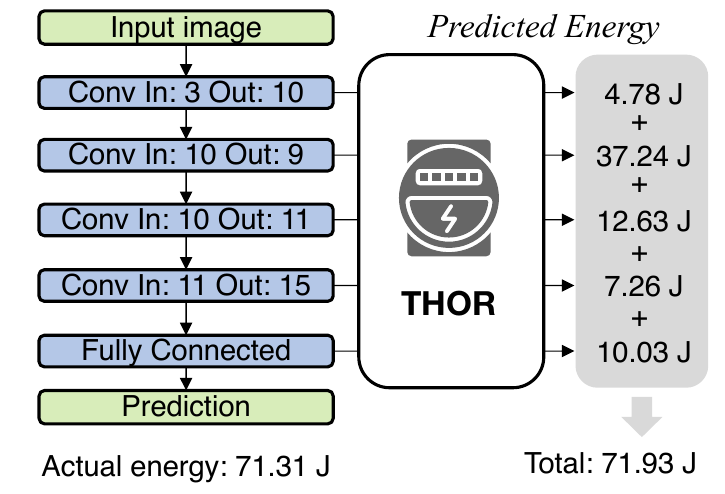}
    \caption{Illustration of the energy estimation process on a 5-layer CNN. In: $a$ and Out: $b$ indicate the input channel $a$ and output channel $b$ for convolutional layers.}
    \label{fig:intro}
\end{figure}

Nascent on-device training is gaining traction due to its data privacy and efficiency benefits, as it circumvents the need to upload data to the cloud. 
Furthermore, it is especially advantageous in situations with unreliable network connectivity or for providing personalized services \cite{xu2022mandheling,cai2021towards,gim2022memory}.
However, given the fact that almost all contemporary mobile devices are powered by battery, their energy consumption is commonly constrained. 
Therefore, deploying \DNN training onto these devices presents a challenge. 
It is a high-power task and requires estimating the overall energy expenditure beforehand to prevent power failure \cite{fraternali2020ember,9869740,RAJU2023109603}.
Addressing these energy constraints is critical for developing sustainable AI, promoting efficient energy use and environmental friendliness.

State-of-the-art solutions may not be sufficiently accurate for estimating \DNN training's energy consumption. 
Evaluating a neural network model's \acfp{FLOP} is a common energy-estimation practice for most models~\cite{Diao2021HeteroFLCA,DBLP:conf/aaai/RappKPH22,EAP}, based on an assumption that the devices’ working status is fixed. 
However, in reality, the energy estimation based on \FLOPs may deviate greatly from the observations. 
Beyond FLOPs, most simulation-based methods are limited to specific hardware and frameworks~\cite{MeiLWSVB22,Rodrigues2018SyNERGYAE,EAP}. 
Developing an accurate and generic energy estimation approach for on-device training presents the following fundamental challenges:

\begin{itemize}[noitemsep,topsep=0pt,parsep=0pt,partopsep=0pt]
    \item \textbf{System Heterogeneity}. 
    Mobile devices have diverse hardware combinations, leading to potential variations in behavior even within the same model. 
    Furthermore, the Lookup Table provided by Neural Architecture Search (NAS) proves impractical for making estimations across a wide range of devices due to this variability \cite{syed2021generalized}.
    
    \item \textbf{Model Diversity}. 
    \DNN models' energy consumption characteristics can vary greatly, inherently resulting in heterogeneous resource demands. Some models may primarily rely on computational power as input scale increases, while others remain largely power-insensitive.
\item \textbf{Runtime Complexity}. It arises from optimization techniques such as kernel fusion and CPU hand-off during model training. 
The unpredictable access patterns from various levels of the memory hierarchy impede the ability to calculate energy usage merely as the sum of data movement and computation.
\end{itemize}

In this paper, we present \sys, a novel and generic method designed to provide precise energy estimates.
It is agnostic to both the running platform and the \DNN model, therefore solves the challenges of system heterogeneity and model diversity.
Based on the observations that layer-associated operators are emitted sequentially and the inter-layer effects are insignificant, we present the layer-wise energy \textit{additivity} and \textit{subtractivity}.
 The total energy consumption of a \DNN is estimated by summing the costs of its layers, with the cost of each layer derived by subtracting the residual layers' costs from the total.
The energy consumption data for individual layers is gathered through layer-wise subtractivity.
Predictive \GP models are fitted with these data, hence avoiding the challenge associated with runtime complexity.
Lastly, the total energy consumption of the entire model can be obtained from summation through layer-wise additivity.
\acf{GP} models are fitted using the observed energy consumption of individual layers.
This fitting process is a one-time endeavor as the resulted models are reusable.
Then, the total energy consumption can be estimated by summing the estimated energy costs of all layers.

We deploy and evaluate \sys across multiple models on five different devices. 
Compared to the current leading approaches, \sys reduces the Mean Absolute Percentage Error (MAPE) by up to 30\%. 
Moreover, we use energy-conscious model pruning to create a leaner architecture with the same performance and 50\% energy consumption, which further verifies the effectiveness and practicality of \sys.
The codes will be publicly released if the paper is accepted.

Our main contributions can be summarized as follows:
\begin{itemize}[noitemsep,topsep=0pt,parsep=0pt,partopsep=0pt]
    \item We study the energy consumption characteristics and find the promising layer-wise \textit{additivity} of energy consumption during \DNN training.
    \item Based on the observations above, we design \sys, a generic method for accurate estimations of energy consumption for training of \DNN models.
    \item We implement \sys and conduct tests on various AI systems.
    The results indicate a reduction in \MAPE by up to 30\%. Moreover, we apply it to guide model pruning and successfully reduce 50\% energy consumption.
\end{itemize}
\section{Background and Motivation}

\subsection{Modeling Deep Learning's Energy Consumption}

\DNN training fundamentally encompasses two primary stages:  \textit{forward} propagation and \textit{backward} propagation. During the process of \textit{forward} propagation, the model takes the input data and calculates the output for each layer. On the other hand, \textit{backward} propagation involves evaluating the difference between the model's output and the target label using a loss function. Subsequently, derivatives are computed in a backward manner, leveraging the chain rule, to guide the update of model parameters.
Each of the aforementioned processes necessitates computational operations, thereby consuming a corresponding amount of energy.

Modeling deep learning energy consumption is a widely studied topic. 
Energy consumption essentially stems from computation and data movement, hence the majority of existing studies primarily focus on estimating this consumption by considering the number of \FLOPs inferred from a \DNN's framework \cite{EAP,chen2016eyeriss,bouzidi2022performance,zhang2021nn,lu2019augur}.
In these methods, the \FLOPS and \FLOPS per watt serve as indicators for computational performance and energy efficiency, respectively. 
we refer to our Appendix Sec. \ref{sec:related} for more thorough exploration of this literature.

\subsection{Heterogeneous Devices}
There are many kinds of devices with different architectures.
Most mobile devices, like smartphones, feature various System-on-Chips as well as integrated \GPUs.
The NVIDIA Jetson platform uses NVIDIA \GPUs, further enhancing its capabilities. 
Desktop platforms are usually equipped with more powerful CPUs and \GPUs.
The energy efficiency ratio of different processors can exhibit orders of magnitude differences~\cite{8782524}.
Data movement is influenced by tensor eviction and prefetch strategies~\cite{peng2020capuchin} and is also determined by the memory hierarchy.
The energy consumption of DRAM can reach up to 200 times that of the register~\cite{sze2017efficient}. 
Even for the same device, the execution efficiency can vary by up to 20 times between kernels with different computational I/O ratios~\cite{monil2020mephesto}.
Therefore, estimating energy consumption using a unified model poses a significant challenge due to the heterogeneous nature of devices.

\begin{figure}[t]
\centering
    \includegraphics[width=\linewidth]{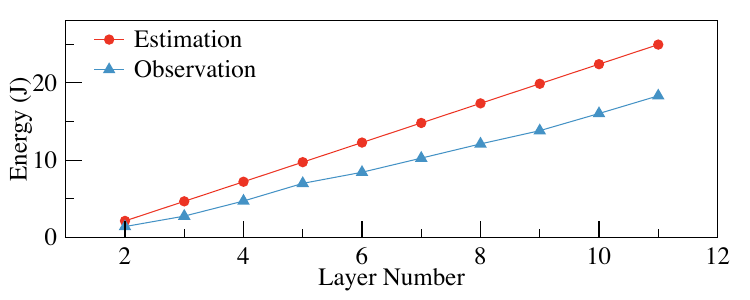}
    \vspace{-0.1in}
    \caption{Energy consumption from NeuralPower estimation and from observation for a CNN.}
    \vspace{-0.1in}
    \label{fig:prev_pred}
\end{figure}

\subsection{Existing Methods' Limitations}
Existing methods fall into three categories~\cite{garcia2019estimation}:

\textbf{Proxy-based methods} estimate energy costs by a model's \FLOPs, parameter size, and number of layers, with \FLOPs-based estimation being the most common~\cite{bakar2022protean,wang2020optimize,han2015learning}.
While adaptable to any model, these methods overlook device heterogeneity and runtime optimizations, assuming stable computation performance and performance-per-watt.
However, when the model structure changes, the system utilization will undergo significant changes. 
The kernel configure tends to launch fewer threads for pruned models~\cite{wang2020time}. 
Upon compilation, the framework generates both forward and backward computational graphs and fuse operations into a single CUDA kernel.
This approach enhances computation for activation functions, optimizers, custom RNN cells, \etc
Some in-place optimizations, such as \texttt{Convolution-BatchNorm-ReLU} fusion, are also implemented and make the execution more like a black box~\cite{Chen2018TVMAA,jia2019taso}.
These factors contribute to inaccuracies in Proxy-based estimations.

\textbf{Simulation-based methods} simulate the full computation and data movement process~\cite{MeiLWSVB22,Rodrigues2018SyNERGYAE,EAP}. 
They require detailed knowledge about the algorithm implementations as well as the energy cont of each hardware component. 
These methods allow for the identification of the hardware that incurs the highest energy consumption and help in locating performance bottlenecks caused by memory stalls. 
However, this type of approach loses its generality and is only applicable to specific devices, models, and \ML frameworks.

\textbf{Architecture-based methods} utilize framework-provided profilers during the inference phase to obtain the execution time for specific layers or kernels, significantly improving the accuracy of energy prediction~\cite{zhang2021nn,cai2017neuralpower,stamoulis2018hyperpower}. 
Nevertheless, this type of approach relies on specific framework and still faces challenges in obtaining energy costs.
As a validation, we extend the forward pass to the whole training process like NeuralPower~\cite{cai2017neuralpower} and adapt it to the training phase as shown in Fig.~\ref{fig:prev_pred}.
We conduct profiling on the operators involved in each of these stages separately and obtain the final energy by summing them up.
The results show that this method tends to overestimate the cost of each layer, which indicates the introduction of systematic biases and verifies our analysis.

\section{\sys's Design}

\subsection{A Bird's-Eye View} 
\label{sec:overview}
\begin{figure}[t]
    \centering
    \includegraphics[width=\linewidth]{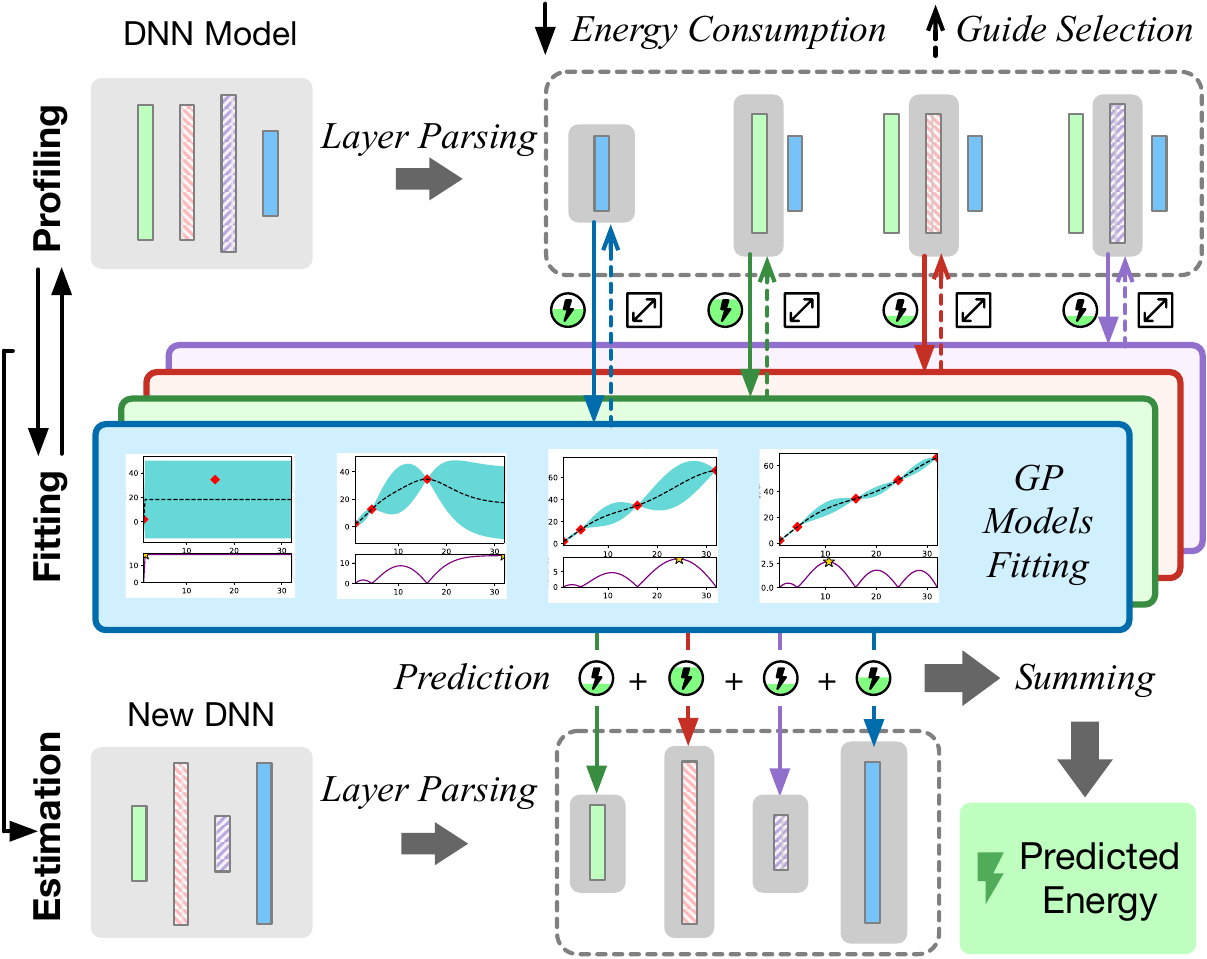}
    \caption{An overview of \sys.}
    \label{fig:overview}
\end{figure}
Fig.~\ref{fig:overview} illustrates an overview of \sys. It broadly contains three processes listed below, where the profiling and the fitting processes are carried out in an iterative manner.

\textbf{Profiling}: In \sys, all layers are parsed into input layer, hidden layer, and output layer.
To estimate the energy consumption of these layers, \sys generates 1-layer, 2-layer, and 3-layer variant NNs with different parameters, respectively. 
Initially, the parameters of variant models are selected as the bound value, and the subsequent parameters are selected and guided by the profiling stage.
Training these variant models provides information about their consumption.

\textbf{Fitting}: 
Based on the \textit{additivity} and \textit{subtractivity} of the energy consumption, \sys separates the model as different layers and fits with \acf{GP} models based on the consumption data from the profiling stage.
Moreover, the \GP models guide the selection of the next profiling point based on predictive uncertainty and utilize the returned data to optimize the prediction.
Therefore, the fitting stage can be seen as an active learning process to some extent.

\textbf{Estimation}: 
Once sufficient energy consumption data are obtained, \GP models are fitted to predict the energy consumption of each layer. 
After that, \sys can obtain the energy estimation by summing the energy from each \GP model based on the additivity of the energy consumption.

The key advantages of our framework are its inherent generality and minimal overhead. By implementing the actual training workflow without the need for an additional operator-level profiler, \sys can be easily integrated into any training framework and device to provide accurate estimations. In essence, \sys offers a highly adaptable and low-overhead solution for a wide range of use cases across diverse computing environments.

\subsection{Profiling}

\textbf{Layer-wise Energy Additivity of DNN.} 
In the energy cost estimation task, the existing method, NeuralPower \cite{cai2017neuralpower}, estimates energy costs by profiling the forward, backward, and update stages. 
However, this approach tends to overestimate costs due to data reuse during the DNN training process. In contrast, we view the DNN model as a combination of different layers and propose that the total energy consumption of the entire DNN can be obtained by summing the costs of each layer, a concept we refer to as layer-wise \textit{additivity}.
 This also implies that the cost of a single layer can be independently assessed by subtracting the cost of the residual layers from the total sum, a concept we call layer-wise \textit{subtractivity}.

To validate this, we train a \CNN model on the MNIST dataset, starting with a rudimentary model including only an input layer and an output (i.e., \texttt{FC}) layer.
After this, we integrate identical \texttt{Conv2d} layers into the existing structure and examine the resulting changes in energy consumption.
Experimental results in Fig.~\ref{fig:prev_pred} empirically illustrate that existing method overestimates the costs.
Moreover, the trends in energy consumption corresponding to the increase in the number of layers.
This underscores that the incremental cost associated with each convolutional layer remains roughly constant, creating a linear trajectory.
This observation suggests that the energy consumption of each layer is additive and the positioning of a layer has no impact on the final outcome.

\textbf{Layer Parsing.}\label{ssec:layer sep}
Based on the presented layer-wise energy additivity pf \DNN, we dissect the network model into sub-components of the three distinct categories: input layers, hidden layers and output layers. 
A \DNN usually contains a single input layer, a single output layer, and multiple hidden layers organized with \textit{Modular Design}, i.e., repeating layers or blocks of layers.
Therefore, we employ a similar strategy to partition them into blocks, where non-parametric layers are consistently grouped together with preceding layers. 
Deduplication is carried out based on the layer type and the associated hyperparameters, \ie, input height and weight, kernel size, and batch size. 
This separation rule can hide the effects of optimization by the framework, therefore, making the estimation more accurate.
The details of our layer characteristics are further described in Appendix Sec. \ref{sec:lc}.

In \sys, layers are characterized by their output channels (for input layers), input channels (for output layers), or both input and output channels (for hidden layers).
For an $n$-layer DNN, the channels are denoted as $C_1$, $C_{n-1}$, and $C_i$ for $i=2, \cdots, n-2$, respectively.
Note that layers with different kernel sizes, steps, and batchsizes are encoded as different layers since their energy cost patterns have a large gap.

\textbf{Profiling Process.} 
In \sys, we firstly profile the output layer, treating it as a complete model for training purposes. 
The energy consumption of this single-layer model is denoted as $E_{\textit{output}}(C_1)$ and we fit a \GP model in the fitting process to predict the energy consumption denoted as $\hat{E}_{\textit{output}}(C_1)$ with varied $C_1$s.

Secondly, the input layer's energy cost can be obtained by subtracting the output layer's estimation result from the total measured cost according to the layer-wise subtractivity:
\def \E(#1,#2){E_{\textit{#1}}(#2)}
\def \Ef(#1,#2){E_{\textit{#1}}^{\textit{fwd}}(#2)}
\def \Eb(#1,#2){E_{\textit{#1}}^{\textit{bwd}}(#2)}
\def \Ew(#1,#2){E_{\textit{#1}}^{\textit{w}}(#2)}
\def \Ey(#1,#2){E_{\textit{#1}}^{\textit{y}}(#2)}
\def \Eu(#1,#2){E_{\textit{#1}}^{\textit{u}}(#2)}
\begin{equation}
\label{eq:input-layer}
\begin{aligned}
    \E(input,C_{2})&=\E(input+output,C_1, C_2)-\hat{E}_\textit{output}(C_1),
\end{aligned}
\end{equation}
and we fit a \GP model in the fitting process to predict the energy consumption denoted as $\hat{E}_{\textit{input}}(C_{2})$ for varied $C_2$s.

Lastly, we assemble the energy consumption of NNs containing an input layer, a single hidden layer, and an output layer with varied $C_1$s and $C_2$s.
Therefore, the energy costs of the hidden layer can be obtained by subtracting estimated costs of others:
\begin{equation}
\label{eq:hidden-layer}
\begin{split}
 E_{\textit{hidden}}(C_1,C_2)
    =\E(model,{C_1,C_2}) -\hat{E}_\textit{input}(C_1)-\hat{E}_\textit{output}(C_2),\\
\end{split}
\end{equation}
and we fit \GP models in the fitting process to predict the energy cost of each kind of hidden layers denoted by $\hat{E}_{\textit{hidden}}(C_1,C_2)$. 

It is noteworthy that the profiling is guided by the fitting process instead of randomly sampling, as shown in detail below. 
It improves the efficiency of the whole framework.

\begin{figure}
  \centering
  \begin{minipage}{0.5\textwidth}
    \centering
    \includegraphics[width=.95\linewidth]{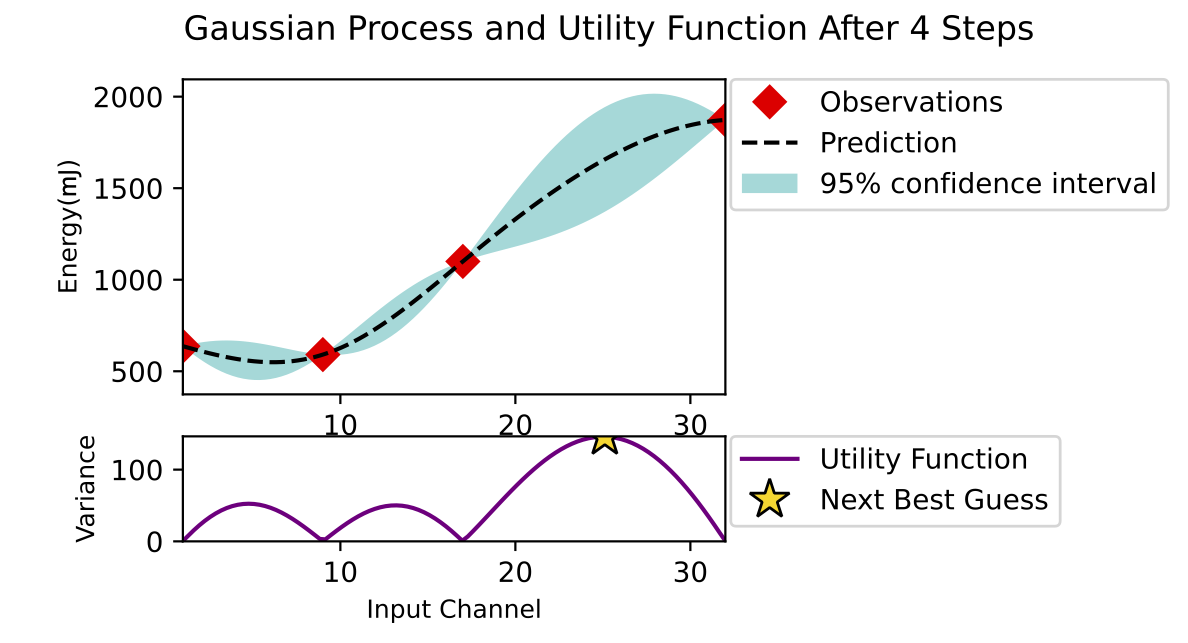}
  \end{minipage}

  \begin{minipage}{0.5\textwidth}
    \centering
    \includegraphics[width=.95\linewidth]{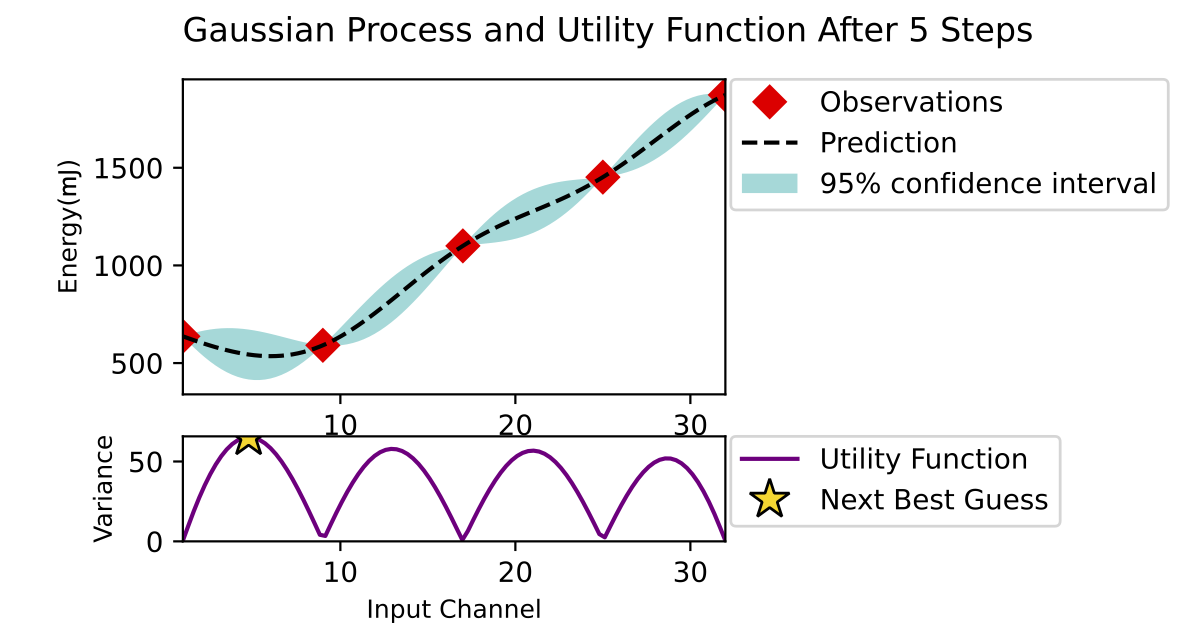}
  \end{minipage}
  \vspace{-0.1in} 
  \caption{\GP after 4 and 5 steps for \texttt{FC} layer on OPPO taking 500 batches of (10, input channel, 28, 28) input.}
  \vspace{-0.1in}
  \label{fig:GP}
\end{figure}
\subsection{Fitting}
\begin{figure}[t]
    \centering
    \includegraphics[width=\linewidth]{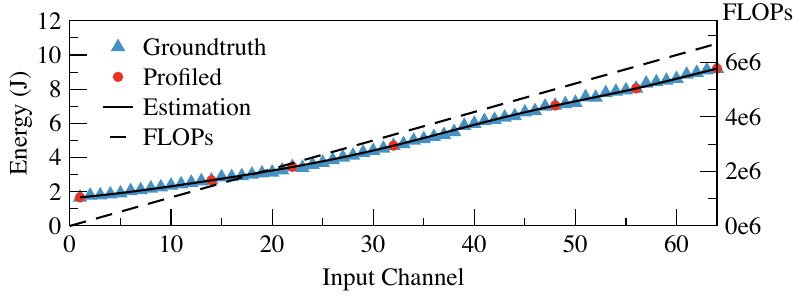}
    \vspace{-0.1in}
    \caption{Energy consumption of a \FC layer taking 500 batches of (4, input channel, 50, 50) input on Xavier.}
    \vspace{-0.1in}
    \label{fig:FC layer}
\end{figure}

\textbf{Gaussian Process Model.} A Gaussian Process (GP) model is a non-parametric, probabilistic model popularly used in machine learning and statistics. 
In \sys, we employ \GP as the model to fit the energy consumption characteristics of each kind of layers. 
GP solely demands the consumption data for fitting, making it align well with the diverse requirements posed by system heterogeneity and model diversity.
It is also capable of handling noise, which is unavoidable due to the potential awakening of background processes in practice.
Moreover, its probabilistic nature provides empirical confidence intervals, allowing adaptive data fitting and assisting in determining termination conditions.

\textbf{Fitting Accurately.} The \GP models are fitted with energy consumption data from the profiling process. 
As illustrated in Fig.~\ref{fig:FC layer}, an increase in the input channel
directly correlates to a linear increase in \FLOPs.
However, the energy consumption does not follow the same linear pattern. 
Hence, utilizing \FLOPs to estimate the energy could result in inaccurate performance predictions, while the \GP model aptly fits
the energy prediction task.

\textbf{Kernel Selection.} In \GP, the prior's covariance is given by a kernel, which can describe how similar two neighbor points are. In \sys, we utilize the Matérn kernel \cite{williams2006gaussian}:
\begin{align}
\begin{aligned}
    k\left(x_{i}, x_{j}\right) = \frac{\left(\frac{\sqrt{2 \nu}}{l} d\left(x_{i}, x_{j}\right)\right)^{\nu}   K_{\nu}\left(\frac{\sqrt{2 \nu}}{l} d\left(x_{i}, x_{j}\right)\right)}{\Gamma(\nu) 2^{\nu-1}}, 
    \end{aligned}
\end{align}
where
$d(\cdot, \cdot)$ is the Euclidean distance, $K_{\nu}(\cdot)$ is a modified Bessel function and $\Gamma(\cdot)$ is the gamma function.
This kernel is generally robust to misspecification of the length-scale parameter \cite{williams2006gaussian}.
It has an additional parameter $\nu$
to control the smoothness of the resulting function. 
Considering the runtime optimization and cache thrashing, we choose $\nu=2.5$ which only requires twice differentiable. 
Experiments provided in Appendix Sec. \ref{sssec:kernel} support the superiority of the Matérn kernel.

\textbf{Guided Profiling.} 
\GP has the unique ability to model the uncertainty hence can guide the selection of the next point in the profiling process.
As our main purpose is to obtain accurate estimation, 
we choose the point with the largest variance to eliminate the uncertainty. 
Fig.~\ref{fig:GP} illustrates how \GP selects the next point and fits the data. 
After fitting the point with the largest variance, the uncertainty is diminished.
Therefore, it can be seen as an instance of active learning to some extent, where the acquisition function guides the selection of the most informative points for fitting in the next step.

\begin{figure}
\centering
\includegraphics[width=\linewidth]{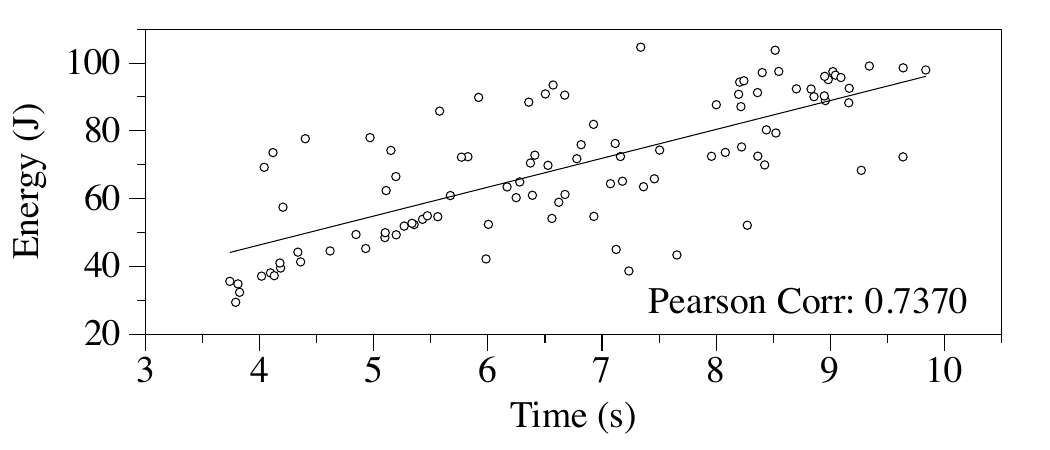}
\vspace{-0.1in}
\caption{The relationship between the time and energy consumption for 5-layer CNN. They have an obvious positive relationship.}
\vspace{-0.1in}
\label{fig:relation}
\end{figure}

However, the real-time energy consumption acquisition is challenging in certain devices like smartphones.
As a solution, we argue there exists a direct correlation between time consumption and energy consumption.
It is verified by experiments conducted on a 5-layer CNN model, as illustrated in Fig. \ref{fig:relation}.
Therefore, we utilize the time uncertainty as a practical surrogate for energy uncertainty to guide profiling.

\textbf{Starting Points and End Condition.} 
To cover the full range of channels, we use the the upper and lower bounds as the starting points. 
On the other hand, as \GP is designed for the continuous situation while the channel number is discrete, the whole process may fall into a dead loop and fails to converge. 
Therefore, we set two end conditions to prevent the worst case: When the number of profiled points exceeds the limits or when the variance is smaller than 5\% of the profiled data, we terminate the profiling and fitting process.

\subsection{Estimation} 
After finishing the profiling and fitting process, we obtain the \GP models for energy cost estimation of each kind of \DNN layers.
According to the layer-wise energy additivity, we dissect any given $n-$layer \DNN model model into three distinct sub-components: input layers, hidden layers and output layers, based on the layer parsing rules described in Sec.~\ref{ssec:layer sep}.
Hence, the total energy consumption of the \DNN can be estimated by summing the estimated energy costs of all layers:
\begin{equation}\label{eq:model}
\begin{split}
    \hat{E}_\textit{model}= & ~\hat{E}_\textit{input}(C_1)+\sum_{i=2}^{n-1}\hat{E}_\textit{hidden}(C_{i-1},C_{i})+\hat{E}_\textit{output}(C_{n-1}),
\end{split}
\end{equation}
where $\hat{E}_{input}$, $\hat{E}_{hidden}$, and $\hat{E}_{output}$ represent \GP{s} to estimate energy costs of input, hidden, and output layers, respectively.

As discussed in Sec.~\ref{sec:intro}, system heterogeneity and model diversity result in significant variations in energy consumption across different systems. 
Therefore, \GP models trained on one device or system cannot be directly transferred to estimate energy consumption on other devices or systems. 
Nevertheless, our \sys framework is generic and broadly applicable, as demonstrated in Sec. \ref{sec:eval}.

\begin{figure}[t]
    \centering
    \includegraphics[width=\linewidth]{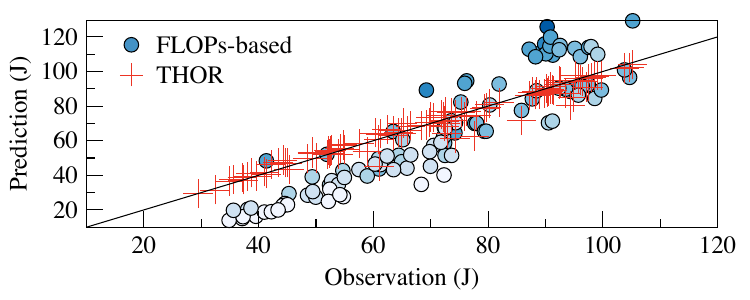}
    \vspace{-0.1in}
    \caption{Estimation results of FLOPs-based method and \sys for a 5-layer CNN.}
    \vspace{-0.1in}
    \label{fig:pred}
\end{figure}
\begin{figure*}[t]
    \centering
    \includegraphics[width=\linewidth]{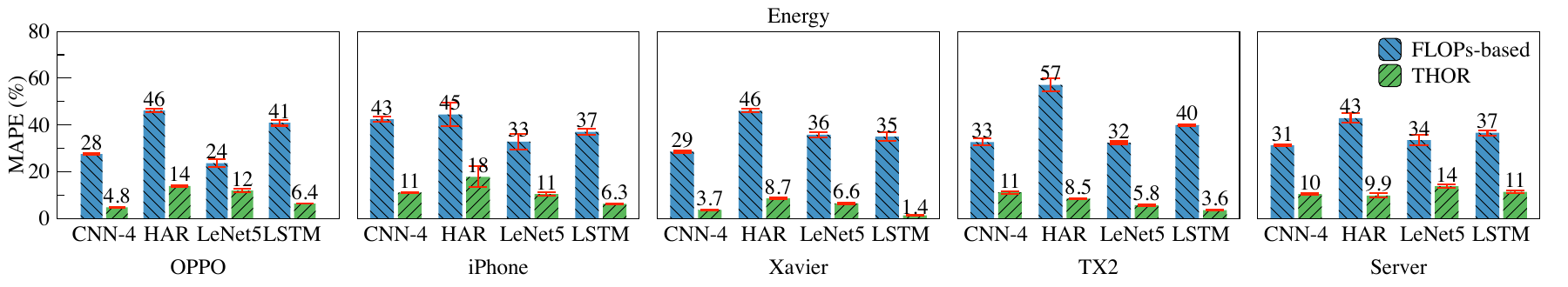}
    \caption{End-to-end energy estimation for five devices.}
    \label{fig:E2E}
\end{figure*}

\section{Evaluation}
\label{sec:eval}

We conduct an evaluation of \sys using multiple representative models on five distinct, realistic devices, including OPPO, iPhone, Xavier, TX2, and Server. 
Details about the data, model architectures, devices, and implementation are provided in Appendix Sec.~\ref{sec:exps}.
\sys significantly decreases the relative errors, with reductions of up to 30\% when compared to previous studies. 
We also apply \sys in a case study focusing on energy-aware model pruning, where it demonstrates a remarkable ability to reduce energy consumption by 50\% without compromising the level of accuracy.

\subsection{End-to-End Estimation Evaluation}
In real-world scenarios, the estimation should be capable of handling unseen models, such as new architectures and parameters. 
To comprehensively evaluate the performance of \sys, we randomly sample the \DNN architectures across channels ranging from 1 to the original channel. 
For the Transformer model, we randomly sample the number of encoder layers and hidden dimensions. 

\textbf{Intuitive Comparison}. To visualize the strengths of \sys compared to the \FLOPs-based method,
we illustrate the energy consumption for a 5-layer \CNN in Fig.~\ref{fig:pred}. 
In the experiment, we generate 100 models with different channels using random sampling.
Alignment with the line indicates an accurate result.
 This reveals that the FLOPs-based method neglects system utilization changes; hence, it tends to overestimate when FLOPs are lower and underestimate otherwise. 
 In contrast, THOR maintains high accuracy across all ranges.

\textbf{Quantitative Comparison}. 
As illustrated in Fig.~\ref{fig:E2E}, our approach has successfully reduced the \MAPE from an average of approximately 40\% to around 10\%.
This significant decrease in \MAPE suggests that our method provides more accurate and stable results. 
The performance of the final estimations varies across devices because of their inherent heterogeneity.
Among the tested devices, the Jetson series, which allows for a fixed frequency, exhibits the most favorable results. 
The estimations for various models on smartphones show a degree of disparity, with some cases, like the HAR model, demonstrating larger errors.
This might be due to the influence of \DVFS policies and power throttling effects. 
On the Server, predictions for different models are relatively consistent, though they have higher error rates compared to other devices.

\textbf{Time Costs.}  
Tab. \ref{tab:tc} presents the time costs for \sys profiling and fitting for various \DNN{s}. 
Most of these tasks are completed within 20 minutes, demonstrating both the efficiency and practical feasibility of the \sys method. 

\begin{table}[t!]
\centering
\caption{Time cost (sec) of profiling and fitting.}
\label{tab:tc}
\begin{threeparttable}
\begin{tabular}{cccccc}
\toprule
 & LeNet5 & 5-layer CNN & HAR & LSTM  \\
\midrule
OPPO   & 694  & 1688 & 2188 & 1615\\
iPhone & 1201 & 1012 & 2446 & 1168 \\
Xavier & 184  & 421  & 740  & 1145  \\
TX2    & 285  & 1211 & 4433 & 422  \\
Server & 235  & 268  & 562  & 436  \\
\bottomrule
\end{tabular}
\end{threeparttable}
\end{table}

\textbf{More Results on Transformers.} 
  \begin{figure}[t]
     \centering
     \includegraphics[width=\linewidth]{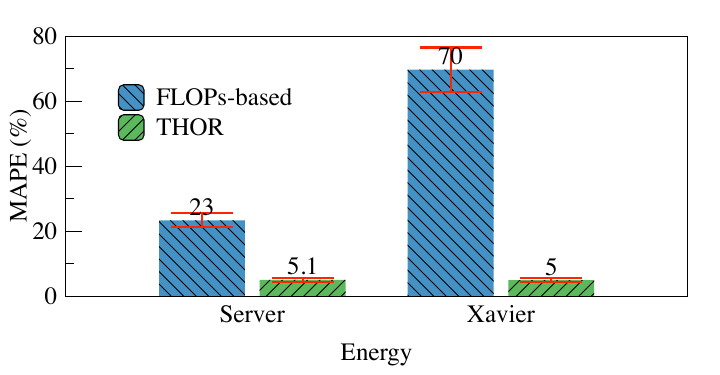}
     \vspace{-0.15in} 
     \caption{Energy estimation of Transformer.}
     \vspace{-0.15in} 
     \label{fig:trans}
 \end{figure}
   \begin{figure}[t]
     \centering
     \includegraphics[width=0.95\linewidth]{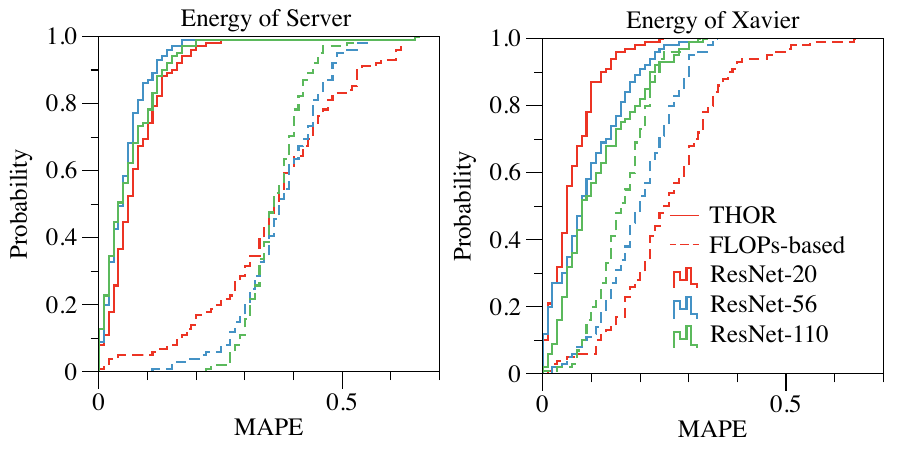}
     \vspace{-0.05in} 
     \caption{ResNet evaluation on Server and Xavier.}
     \vspace{-0.15in} 
     \label{fig:resnet}
 \end{figure}
  \begin{figure*}[t]
    \centering
    \includegraphics[width=\linewidth]{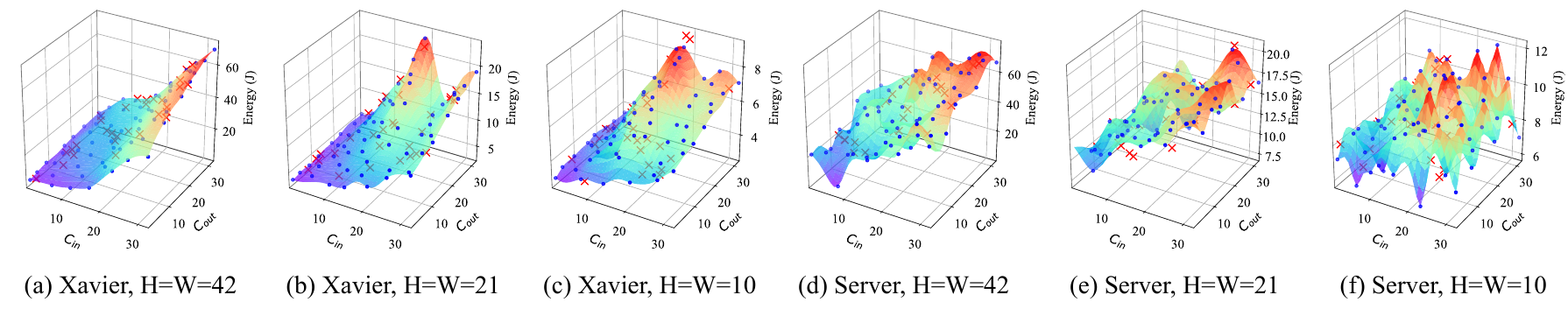}
    \vspace{-0.15in}
    \caption{Profiled v.s. estimated energy for Xavier and Server (The \blue{blue {\Huge $\cdot$}} represents the profiled samples, the \red{red $\times$} represents the test samples, and the surface represents the estimated data).}
    \vspace{-0.15in} 
    \label{fig:conv xs}
\end{figure*}
\begin{figure}[t]
    \centering
    \includegraphics[width=\linewidth]{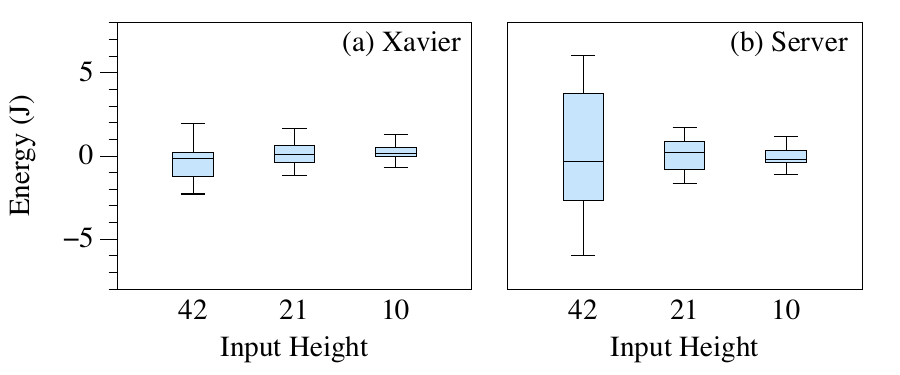}
    \vspace{-0.15in}
    \caption{Differences of energy estimation and observation for Xavier and Server.}
    \vspace{-0.15in} 
    \label{fig:difference}
\end{figure}
To further demonstrate the superiority of \sys, we present the energy estimation results for the Transformer architecture using both \sys and \FLOPs-based methods in Fig. \ref{fig:trans}.
Due to memory restrictions, only the Xavier and Server platforms can fully execute the Transformer model.
\sys consistently outperforms \FLOPs-based methods in energy estimation of the Transformer, further underscoring the effectiveness and generality of \sys.

 \textbf{CDF Plot of ResNets.} To further assess the scalability of \sys, we conduct experiments on the ResNet family of models. 
Due to memory constraints, we can only execute these experiments on the Xavier and Server devices. 
During the profiling phase, we sample ten different layers based on existing rules in Sec.~\ref{ssec:layer sep}. 
The resultant \CDF plot is shown in Fig.~\ref{fig:resnet}.
A step curve closer to the top-left corner indicates higher prediction accuracy. 
The results from both the Xavier and Server devices show improved performance compared to the \FLOPs-based approach. 
Moreover, as the number of layers increases, the prediction accuracy does not appear to decrease.

\subsection{Layer Characteristics}\label{sec:expa}
We present the energy consumption of \texttt{Conv2d} layers in the 5-layer CNN model, as they account for the majority of computational costs. 
The relevant sampling and estimation results across different devices are illustrated in Fig.~\ref{fig:conv xs}, where $H$ and $W$ represent height and width, respectively, and $C_{in}$ and $C_{out}$ denote the input and output channels. 
The batch size is set as $10$.
We use additional random points to test the estimation results, and the differences are shown in Fig.~\ref{fig:difference}.
The energy consumption exhibits a non-linear variation with respect to the input and output channels. 
In these scenarios, the energy consumption is relatively smooth when $H$ and $W$ are large. 
For Xavier, the energy does not initially change significantly with the input channel and remains at a plateau. 
However, after $C_{in}$ exceeds 20, there is a noticeable increase. 
When $H$ and $W$ are at their smallest values, the energy of Server fluctuates around the average, while Xavier demonstrates a distinct ridge in the middle. 
The power consumption characteristics of the Server are more complex, leading to poorer estimation results. 
Although the relative error may be small when the input size is large, it may still contribute the most to the final absolute error. 
We also compare the energy efficiency of these devices, with Xavier and Server exhibiting similar performance for these layers.
Once we have identified these characteristics, we can compute the corresponding gradients. 
This allows us to more effectively guide the model pruning and architecture search, while avoiding extreme outcomes like increased energy consumption after pruning.
\subsection{Case Study: Energy-Aware Pruning}
\begin{figure}[t]
    \centering
    \includegraphics[width=\linewidth]{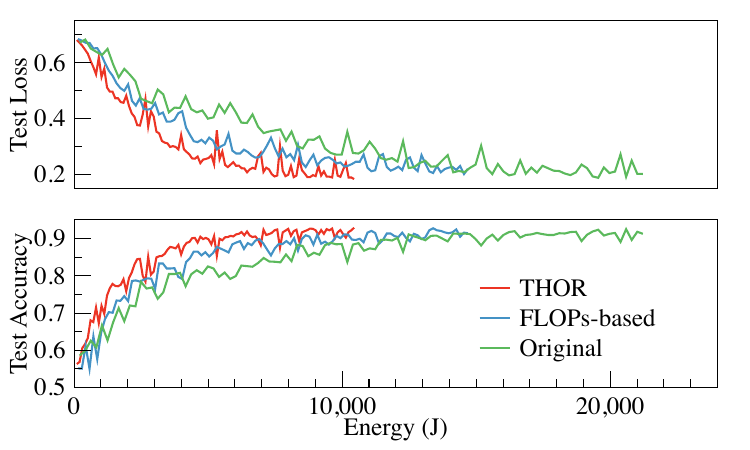}
    \vspace{-0.1in}
    \caption{Test loss and accuracy for different models.}
    \vspace{-0.1in} 
    \label{fig:acc}
\end{figure}

To verify that \sys meets the objective of being easy to combine with other methods, we undertake the task of gender identification from the real-world CelebA dataset on Xavier. 
The original model's total energy consumption stands at approximately 20,000J over 2,000 iterations. 
We simulate an energy-constrained environment where the available energy for consumption is curtailed to $50\%$ of its initial level.
To achieve this target, we apply random pruning as suggested by Li et al.~\cite{li2022revisiting}. 
We then incorporate the \sys and \FLOPs-based methods to guide the process until the energy consumption per iteration drops to $50\%$ of its original value.
As illustrated in Fig.~\ref{fig:acc}, both methods effectively reduce the total energy consumption. 
However, only the \sys approach manages to keep the total energy consumption below the allotted budget ($49.2\%$).
The \sys method demonstrates superior performance by effectively keeping the total energy consumption within the defined budget.
This result ensures efficient resource allocation and highlights the potential of \sys as a revolutionary tool in energy-constrained scenarios. Consequently, it presents \sys as a more viable solution compared to the traditional \FLOPs-based method.

\section{Conclusion}
\label{sec:conclusion}
This paper proposes \sys, a generic method to estimate the energy consumption of \DNN training.
\GP is used to fit layerwise consumptions, then the end-to-end estimation can be obtained by summing the energy consumption predictions of each layer based on the presented layer-wise energy additivity.
We demonstrate the effectiveness of \sys for different \DNN architectures and real-world devices. 
Moreover, \sys can be easily integrated into existing training frameworks to guide energy-aware job scheduling.

\clearpage

\bibliographystyle{named}
\bibliography{refs}

\clearpage
\appendix
\renewcommand\thesection{A\arabic{section}}
\renewcommand{\thetable}{A\arabic{table}}
\renewcommand{\thefigure}{A\arabic{figure}}
\renewcommand{\thealgorithm}{A\arabic{algorithm}}

\section*{Appendix}

\section{More discussions on the Necessity of Accurate Estimation}
Mobile devices are often constrained by the limited power supply and computational resources compared to cloud-based devices.
They may be battery-powered or even self-powered by harvesting their required energy from the environment. 
Therefore, the task scheduling and energy management modules within these devices rely heavily on accurate energy estimations to ensure optimal and sustainable performance.
Furthermore, given that these units typically execute energy-sensitive tasks, there is an inherent necessity for precise energy estimation.
Inaccurate energy estimates can lead to the misallocation of tasks and, in severe instances, precipitate power failures \cite{RAJU2023109603,9869740,9984803}.

Many researchers make an oversimplified assumption that unrealistically represents the problem~\cite{tran2019federated,nguyen2020efficient,yang2020energy,li2021talk,rouhani2016delight}. 
These prior works typically compute time by dividing the total number of clock cycles required for computation by the clock frequency, which parallels the FLOPs-based method when frequency remains constant. 
Moreover, these studies often assume that the power remains constant during the experiments, regardless of varying factors, such as incremental batch sizes, which does not reflect real-world conditions.
For example, AccelWattch \cite{kandiah2021accelwattch} developed a model for GPU power that considers the total number of active Streaming Processors, thread number, and thread utilization. 
These factors can significantly vary across different kernels, leading to diverse energy consumption and latency trends.

\section{Related Work}
\label{sec:related}

\textbf{Energy Estimation for Deep Learning.} Some research has analyzed both the energy consumption and the environmental concerns associated with training on servers or clusters~\cite{strubell2020energy,gupta2021chasing}. They can be studied by a trace-driven method of large-scale analysis of real-world jobs~\cite{hu2021characterization}.
However, at the device level, there is no such systematic research. 
On mobile devices, there are mainly two types of energy consumption measurement: software-based methods and hardware-based methods. Software-based methods are provided by the OS, which can obtain more fine-grained data, while hardware-based methods require external devices but offer higher accuracy~\cite{ahmad2015review}. 
In our experiments, we use software-based methods for systems that provide corresponding interfaces, and hardware-based methods otherwise. 
Since the latency of the corresponding operations in \DNN can be too short to sample, existing methods only focus on the whole model level or use peak power~\cite{lu2019augur,bouzidi2022performance,cai2021towards}.
As for energy estimation, 
most methods use \FLOPs as proxies for energy consumption, and others use these as input features for a regressor to make estimations~\cite{liudarts,lipruning}. 
These methods are not accurate enough due to heterogeneous hardware specifications and runtime performance of the devices.

 \textbf{Energy-Aware Inference and Training.} 
%
When the energy required for inference can be predicted, servers can design a better model architecture based on this information. 
EAP and ChamNet achieved higher accuracy than handcrafted and automatically designed architectures~\cite{dai2019chamnet,EAP}.
Given the limited energy and capabilities of individual devices, it is essential to predict the training time and energy consumption for each device.
Recently, Zeus \cite{you2023zeus} was proposed to achieve a better tradeoff between model performance and energy consumption. However, it does not provide a clear method to estimate the energy consumption in model training.
Moreover, existing approaches tend to oversimplify the estimation and management of energy consumption. 
For instance, they assume constant power levels or equate total computational load solely to the \FLOPs~\cite{tran2019federated,nguyen2020efficient,yang2020energy,li2021talk}. 
The inaccurate results may lead to degraded scheduling performance.
Our method can further guide model pruning accurately, similar to tailorFL~\cite{deng2022tailorfl}.

%
%
%

%
%

%

\section{Layer Characteristics} \label{sec:lc}

\ac{DNN}s are neural networks that have multiple layers between the input and output layers. The general architecture pattern of a DNN consists of the following components:

\noindent \textbf{Input Layer}: This layer receives the input data or features. The number of neurons in this layer must align with the dimensions of the input data. It can also be an embedding layer, which is commonly used in natural language processing tasks to convert discrete categorical input data (\eg words or tokens) into continuous vectors. The embedding layer tends to have a large parameter size and can be initialized or learned from scratch during the training process. 

\noindent \textbf{Hidden Layers}: These layers are responsible for transforming the input data and extracting complex features. By stacking layers, \DNN can learn hierarchical features from the input data. Lower layers tend to capture more simple and local features (\eg, edges and textures in images), while deeper layers can learn more complex and abstract features that are combinations of the lower-level features. This ability to learn hierarchical representations makes deep networks more expressive and capable of solving complex tasks~\cite{DBLP:conf/nips/QiYSG17}. 

In the design of \ac{DNN}, the hidden layers tend to be organized with \textbf{Modular Design}: Repeating layers or blocks of layers can make the architecture more modular and easier to design. 
This is particularly useful when designing networks with varying depths for different tasks or computational budgets. 
For example, in ResNet, the same residual block can be repeated a different number of times to create networks with different depths (\eg, ResNet-20, ResNet-56, ResNet-110, \etc). 
The hidden layers have the largest channel space, which makes them hard to profile.
\par \noindent \textbf{Output Layer}: This is the final layer (\texttt{\FC} layer) in the network that produces the desired output, such as class probabilities for a classification task or predicted values for a regression task. 
Conceptually, this layer can be represented as a one-layer model that completes the training process. 
The number of neurons corresponds to the dimensions of the intended output, a constant that is specific to each job.

\section{Discussions and Limitations} \label{sec:dal}
\noindent \textbf{Parallel Execution}: Our estimation method is based on the assumption that layers are executed in a sequential order. However, some model architectures, such as GoogleNet and SqueezeNet, have independent layers or parallel layers, resulting in more complex predictions. 

\noindent \textbf{Framework Version and Workload}: We test the performance of our model on different versions of the same framework and found that accuracy may decrease when the version changes. For example, we profile for Tensorflow.js@3.19.0 but test on Tensorflow.js@4.7.0, the \MAPE on OPPO increases from 8\% to 11\%. Regular re-training may be necessary, and the corresponding threshold can be determined based on \GP's variance. In our experiments, all devices are running only training process, and the performance may be affected if other tasks are executing simultaneously.

\noindent \textbf{Transferability}: As we discuss in Sec. 1, system heterogeneity and model diversity make energy consumption of multiple models hugely on different system differ hugely. Even though \sys is general and can be used widely, \sys models trained on one scenario cannot be directly transferred to other scenarios.

\section{Experimental Details} \label{sec:exps}

\subsection{Experimental Settings} 

\textbf{Test DNN Models.} To evaluate the effectiveness of our system, we test \sys on five different models. These models are:
\begin{itemize}[leftmargin=*]
    \item LeNet-5 \cite{lecun1998gradient}: A pioneering convolutional neural network designed primarily for handwritten digit recognition.
    \item ResNet \cite{he2016deep}: 
    A deep neural network architecture that employs skip connections, addressing the vanishing gradient problem in deep networks.
    \item 5-layer CNN: A 5-layer CNN comprising four \texttt{Conv2D+BatchNorm+MaxPooling}  layers and a subsequent \texttt{FC} layer.
    \item LSTM: A network with an embedding layer, two stacked LSTM layers (Units=128,128) interspersed with Dropout layers, and a final FC layer (Units set to vocabulary size).
    \item Transformer \cite{vaswani2017attention}: An architecture revolutionized natural language processing tasks by using attention mechanisms.
\end{itemize}

These models have been widely utilized across a range of tasks, such as image classification, word prediction, and human activity recognition. 
To ensure broad applicability of our findings and mitigate potential biases, we utilized random data for our experiments unless otherwise specified.
The shapes of the data follow the popularly used datasets, including FEMNIST \cite{caldas2018leaf}, CelebA \cite{liu2015faceattributes}, ImageNet \cite{deng2009imagenet}, MotionSense \cite{Malekzadeh:2019:MSD:3302505.3310068}, etc.
 \begin{figure}[t]
     \centering
     \includegraphics[width=\linewidth]{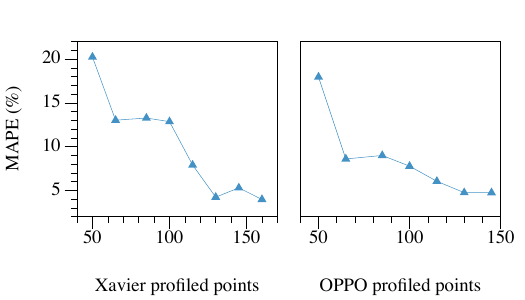}
     \vspace{-0.2in}
     \caption{Correlation between profiled points and the MAPE for energy and time for OPPO and Xavier.}
     \vspace{-0.2in} 
     \label{fig:points}
 \end{figure}
  \begin{figure}[t]
     \centering
     \includegraphics[width=\linewidth]{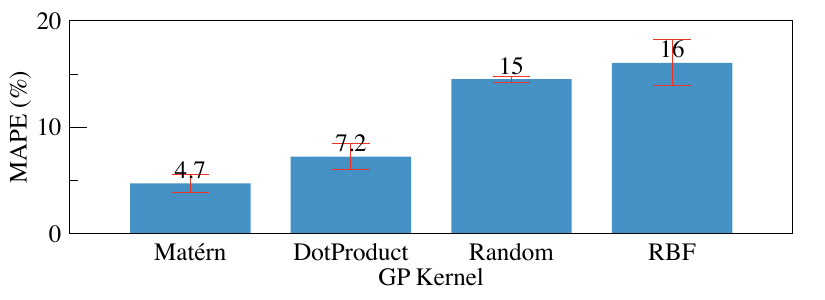}
     \vspace{-0.25in}
     \caption{Energy estimation performance for different \GP kernels  (Random: the points are fitted with Matérn kernel and selected by random sampling).}
     \vspace{-0.2in} 
     \label{fig:kernel}
 \end{figure}

  \textbf{Comparison Baseline.} 
As a representative proxy-based method, \FLOPs is the most popular estimation metric for energy consumption. Therefore, we use \FLOPs as the input to fit a Linear Regression Model to obtain the energy consumption estimation.
The \FLOPs are obtained using the \texttt{torchinfo} module, which can summarize the \FLOPs and memory usage. 
As our target is to build a generic energy consumption approach, existing simulation-based and architecture-based methods cannot be directly used as these methods are only applicable to specific devices, models, and ML frameworks.

 \textbf{Metrics.} We evaluate the estimation performance using the Mean Absolute Percentage Error (MAPE): 
\begin{equation}
    MAPE = \frac{1}{n} \sum\frac{|(Actual_i - Estimated_i)|}{|Actual_i|} \times 100\%.
\end{equation}
It measures the average percentage difference between the actual values and the estimated values, providing an easy-to-interpret and scale-independent measure of forecast error. 
To ensure the stability and reliability of the data, we randomly sample 100 structures for each model and calculate the \MAPE. 
The experimental process mentioned above is repeated three times, and we report the mean value and the standard error. 
For ResNet models, we also plot the \CDF, which describes the error distribution.

 \subsection{System Implementation}
\textbf{Decoupled Architecture.} 
As outlined in Sec.~\ref{sec:overview}, our \sys framework comprises three stages: the profiling stage, fitting stage, and estimation stage. 
These are implemented as distinct Python programs.
The profiling stage is executed via a \texttt{client} program, while the fitting stage is conducted through a \texttt{server} program. The \texttt{client} program operates on the device to estimate energy consumption. 
Interestingly, the server can also operate on separate devices, facilitated by network socket implementation to ensure communication between them. 
The only requirement for running the server is the installation of the corresponding version of the \GP library.
Similarly, the third stage, the estimation stage, is implemented as an individual program. 
This component has the flexibility to run on various devices as an offline process since it only requires the previously recorded energy and fitting data.

 \textbf{Frameworks \& Devices.} The \DNN frameworks can provide a high-level abstraction for model training. In our experiment, we use PyTorch for NVIDIA \GPUs and Tensorflow.js for other devices to train the models. 
PyTorch is based on underlying libraries like cuDNN/cuBLAS and \CUDA API. 
Tensorflow.js is implemented with WebGL, which is a JavaScript API for rendering high-performance graphics on browsers. 
Although the performance of TensorFlow.js may be inferior, it offers the advantage of being platform-agnostic and device-agnostic. 
We conduct our experiments on heterogeneous devices that are commonly used, including smart phones, development boards and a Windows server. 
Specifications of each device are listed in Tab.~\ref{tab:devices}. 

\begin{table*}[t]
\centering
\caption{ Summary of the heterogeneous devices.}
\label{tab:devices}
\resizebox{\textwidth}{!}{
\begin{tabular}{lllll}
\toprule
\textbf{Type}                                                                     & \textbf{Device Name}    &\textbf{Short Form}       & \textbf{Processor}                                                                                                                        & \textbf{GPU}                                                                                                     \\ \midrule
\multirow{2}{*}{\begin{tabular}[c]{@{}l@{}}Smart\\ Phones\end{tabular}} & OPPO Reno6 Pro+ &OPPO & Qualcomm SM8250-AC Snapdragon 870                                                                                                & Adreno 650                                                                                              \\ \cline{2-5} 
                                                                         & iPhone13  &iPhone       & Apple A15 Bionic                                                                                                                 & Apple GPU (4-core graphics)                                                                             \\ \midrule
\multirow{3}{*}{Boards}                                                  & Jetson Xavier NX &Xavier& 6-core NVIDIA Carmel Arm®v8.2 64-bit CPU                                                                                         & \begin{tabular}[c]{@{}l@{}}384-core NVIDIA Volta™ architecture GPU \\ with 48 Tensor Cores\end{tabular} \\ \cline{2-5} 
                                                                         & Jetson TX2  &TX2     & \begin{tabular}[c]{@{}l@{}}Dual-core NVIDIA Denver™ 2 64-bit CPU \\ and quad-core Arm® Cortex®-A57 MPCore processor\end{tabular} & 256-core NVIDIA Pascal™ architecture GPU                                                                \\ \midrule
Server                                                                   & Windows Server &Server  & Intel i9-13900K                                                                                                                  & NVIDIA GeForce RTX 4090                                                                                \\ \bottomrule
\end{tabular}
}
\end{table*}
\textbf{Energy Consumption Acquisition.} 
\begin{figure}[t]
    \includegraphics[width=\linewidth]{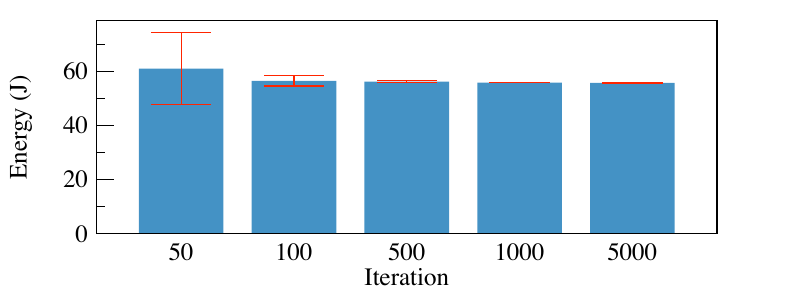}
    \vspace{-0.1in}
    \caption{Normalized energy for LeNet on Xavier with different iterations.}
    \vspace{-0.1in}
    \label{fig:resource profile}
\end{figure}
The energy consumption from \DNN training can be obtained by recording the difference between the measured energy consumption during training and the standby energy consumption. The energy is calculated as follows:
\begin{equation}
    E = \int_{0}^{T} P(t) dt \approx \sum_{i=0}^{n-1} P(t_i) \Delta t = \bar P*T,
\end{equation}
where $E$ is the energy consumption, $P(t)$ represents the power at time $t$, $\Delta t$ is the sampling interval, and $T$ is the total time. 
The total cost of energy can be obtained by multiplying the number of iterations by the cost per iteration. 
As a single iteration is too short to capture its characteristics, we profile specific iterations and average them to get the true costs.
As depicted in Fig.~\ref{fig:resource profile}, we conduct experiments of different iterations and normalize the energy consumption to 1000 iterations.  
Insufficient profiling iterations may lead to instability and discrepancies in the results.
In our experiments, we select 500 as the representative number for profiling iterations.
The final energy consumption can be calculated by multiplying the total iterations by the normalized cost per iteration.

The energy consumption of smartphones is measured using an external power monitor, the Chargerlab POWER-Z KT002~\cite{chargerlab}. 
This versatile mobile device USB tester is capable of reading power bus data at a frequency of 10Hz without exerting any additional load on the test device.
NVIDIA equips the Jetson Module with comparable power monitors like the INA3221. 
The system's power draw can be accessed through the Linux \textit{sysfs} interface under \textit{/sys/bus/i2c/drivers/ina-}\textit{3221x/}. 
We assess the total power consumption of the entire device by reading the output from \textit{in\_power0\_input}. 
In our experiments, we observe that setting the sampling interval to $1$ms results in performance degradation, leading to an approximate 10\% slowdown. 
Thus, after testing various intervals, we select $100$ms as the optimal minimum interval to prevent the detrimental effects on performance.
For the high-end server, we use \textit{nvidia-smi} to collect real-time power data at its maximum frequency of approximately 50Hz. 
Before conducting any experiments, we close all background applications on the device to stabilize energy consumption as far as possible.

 \section{Sensitivity Analysis and Ablation Test} \label{sec:expse}
 \subsection{Varying Profiling Points}

We investigate the influence of varying the number of sampled data points on the performance of the 5-layer CNN model. The detailed examination of the experimental results is depicted in Figure~\ref{fig:points}.
In general, an increase in the number of sampled data points tends to yield enhanced performance for the prediction result of the \GP. This can be attributed to the fact that a larger sample size provides a more comprehensive representation of the underlying data distribution, thereby leading to more accurate and reliable  predictions.
However, the benefits associated with increasing the sample size exhibit diminishing returns as the number of data points reaches a threshold. Additional data points contribute only marginally to the improvement of the \GP's understanding of the data distribution, it may even overfit the samples.
Furthermore, it is crucial to consider the potential drawbacks of excessive sampling, which may lead to significant time and energy consumption during the data profiling phase since the computational cost grows linearly. Consequently, we try to strike a balance between the number of sampled data points and the overall performance of the \GP, taking into account the convergence and end conditions.

 \subsection{Selection of Different \GP Kernels and Efficiency of \GP}\label{sssec:kernel}
In this subsection, we present an analysis focusing on the comparative performance of different kernel functions used in \GP. Specifically, we evaluate the DotProduct kernel, the \RBF kernel, and the Matérn kernel. The analysis aims to find the most suitable kernel for our framework.
The DotProduct kernel is a non-stationary kernel that is particularly adept at modeling linear relationships between variables. Its mathematical representation is given by:
\begin{equation}
    k(x_i, x_j) = x_i \cdot x_j + \sigma_0^2,
\end{equation}
where $x_i$ and $x_j$ are input vectors, $\sigma_0$ is a scale parameter, and $d$ is the degree of the polynomial.
On the other hand, the \RBF kernel is a widely-used kernel function capable of modeling complex, non-linear relationships. However, its performance is sensitive to the choice of hyperparameters, which may result in overfitting in some cases. The \RBF kernel is defined as:
\begin{equation}
    k(x_i, x_j) = \exp\left(-\frac{||x_i - x_j||^2}{2\sigma^2}\right),
\end{equation}
where $\sigma$ is a length-scale parameter.
As demonstrated in Figure~\ref{fig:kernel}, the Matérn kernel outperforms both the DotProduct and \RBF kernels in our experiments. The DotProduct kernel ranks second in terms of performance, while the \RBF kernel exhibits the poorest results.
To further illustrate the efficiency of \GP, we conduct additional experiments using random sampling to generate data points, which are subsequently fit using the Matérn kernel. The results reveal that \GP achieves higher accuracy in comparison to random sampling. This observation suggests that \GP is more effective at exploring the search space and identifying the optimal solution.

\clearpage
\appendix
\renewcommand\thesection{A\arabic{section}}
\renewcommand{\thetable}{A\arabic{table}}
\renewcommand{\thefigure}{A\arabic{figure}}
\renewcommand{\thealgorithm}{A\arabic{algorithm}}

\end{document}